\documentclass{article}

\usepackage{microtype}
\usepackage{graphicx}
\usepackage{subcaption}
\usepackage{booktabs} 

\usepackage{hyperref}

\usepackage[preprint]{icml2026}

\usepackage{amsmath}
\usepackage{amssymb}
\usepackage{mathtools}
\usepackage{amsthm}

\usepackage[capitalize,noabbrev]{cleveref}

\theoremstyle{plain}

\theoremstyle{definition}

\theoremstyle{remark}

\usepackage[textsize=tiny]{todonotes}

\usepackage[nolist,nohyperlinks]{acronym}
\usepackage[inline]{enumitem}
\usepackage[dvipsnames]{xcolor}
\usepackage{multirow}

\newcommand{\ourmethod}{SATTS}

\newcommand{\textbfp}[1]{\vspace{0.5em}\noindent\textbf{#1.}}

\begin{acronym}[LONGEST]
  \acro{CV}{Computer Vision}
\end{acronym}
\begin{acronym}[LONGEST]
  \acro{NLP}{Natural Language Processing}
\end{acronym}
\begin{acronym}[LONGEST]
  \acro{PDE}{Partial Differential Equation}
\end{acronym}
\begin{acronym}[LONGEST]
  \acro{CFD}{Computational Fluid Dynamics}
\end{acronym}
\begin{acronym}[LONGEST]
  \acro{DEM}{Discrete Element Method}
\end{acronym}
\begin{acronym}[LONGEST]
  \acro{SPH}{Smoothed Particle Hydrodynamics}
\end{acronym}
\begin{acronym}[LONGEST]
  \acro{UDA}{Unsupervised Domain Adaptation}
\end{acronym}
\begin{acronym}[LONGEST]
  \acro{DA}{Domain Adaptation}
\end{acronym}
\begin{acronym}[LONGEST]
  \acro{FDM}{Finite Difference Method}
\end{acronym}
\begin{acronym}[LONGEST]
  \acro{FEM}{Finite Element Method}
\end{acronym}
\begin{acronym}[LONGEST]
  \acro{FE}{Finite Element}
\end{acronym}
\begin{acronym}[LONGEST]
  \acro{FVM}{Finite Volume Method}
\end{acronym}
\begin{acronym}[LONGEST]
  \acro{UPT}{Universal Physics Transformer}
\end{acronym}
\begin{acronym}[LONGEST]
  \acro{MSE}{Mean Squared Error}
\end{acronym}
\begin{acronym}[LONGEST]
  \acro{RMSE}{Root Mean Squared Error}
\end{acronym}
\begin{acronym}[LONGEST]
  \acro{IWV}{Importance Weighted Validation}
\end{acronym}
\begin{acronym}[LONGEST]
  \acro{DEV}{Deep Embedded Validation}
\end{acronym}
\begin{acronym}[LONGEST]
  \acro{GNN}{Graph Neural Network}
\end{acronym}
\begin{acronym}[LONGEST]
  \acro{PEEQ}{Equivalent Plastic Strain}
\end{acronym}

\begin{acronym}[LONGEST]
  \acro{TTA}{Test-Time Adaptation}
\end{acronym}
\begin{acronym}[LONGEST]
  \acro{SSA}{Significant-Subspace Alignment}
\end{acronym}
\begin{acronym}[LONGEST]
  \acro{MAE}{Mean Absolute Error}
\end{acronym}
\begin{acronym}[LONGEST]
  \acro{MMD}{Maximum Mean Discrepancy}
\end{acronym}
\begin{acronym}[LONGEST]
  \acro{DPP}{Determinantal Point Processes}
\end{acronym}
\begin{acronym}[LONGEST]
  \acro{DDPM}{Denoising Diffusion Probabilistic Models}
\end{acronym}

\icmltitlerunning{Stable Adaptation at Test-Time for Simulation (\ourmethod)}

\begin{document}

\twocolumn[

  \icmltitle{Stabilizing Test-Time Adaptation of High-Dimensional Simulation \\ Surrogates via D-Optimal Statistics}



  \icmlsetsymbol{equal}{*}

  \begin{icmlauthorlist}
    \icmlauthor{Anna Zimmel}{unv}
    \icmlauthor{Paul Setinek}{unv}
    \icmlauthor{Gianluca Galletti}{unv}
    \icmlauthor{Johannes Brandstetter}{unv,comp}
    \icmlauthor{Werner Zellinger}{unv}
  \end{icmlauthorlist}

  \icmlaffiliation{unv}{ELLIS Unit, LIT AI Lab, Institute for Machine Learning, JKU Linz, Austria}
  \icmlaffiliation{comp}{Emmi AI GmbH, Linz, Austria}

  \icmlcorrespondingauthor{Anna Zimmel}{zimmel@ml.jku.at}

  \icmlkeywords{Machine Learning}

  \vskip 0.3in
]



\printAffiliationsAndNotice{}  

\begin{abstract}

Machine learning surrogates are increasingly used in engineering to accelerate costly simulations, yet distribution shifts between training and deployment often cause severe performance degradation (e.g., unseen geometries or configurations).
Test-Time Adaptation (TTA) can mitigate such shifts, but existing methods are largely developed for lower-dimensional classification with structured outputs and visually aligned input-output relationships, making them unstable for the high-dimensional, unstructured and regression problems common in simulation.
We address this challenge by proposing a TTA framework based on storing maximally informative (D-optimal) statistics, which jointly enables stable adaptation and principled parameter selection at test time. When applied to pretrained simulation surrogates, our method yields up to 7\% out-of-distribution improvements at negligible computational cost. To the best of our knowledge, this is the first systematic demonstration of effective TTA for high-dimensional simulation regression and generative design optimization, validated on the SIMSHIFT and EngiBench benchmarks.

\end{abstract}

\section{Introduction}
\label{intro}


Neural surrogates have become powerful tools for accelerating \ac{PDE} simulations across engineering and science. 
They perform well when test conditions match the training data, but performance often drops on unseen configurations (geometry, material types, structural dimensions, desired and physical parameters), i.e., when the data distribution shifts.
This challenge gets more pronounced in industrial simulation and design optimization, where configurations can vary widely across iterations and frequently extend beyond the ranges known a priori, at data generation and training time. 
In many cases, only large pre-trained surrogate models are available, making full retraining costly or impractical. Moreover, access to original training data may be limited due to portability or proprietary constraints, highlighting the need for model- and task-agnostic approaches that enable zero-shot adaptation and automated model selection during design optimization.

The problem of tackling distribution shifts~\citep{quinonero2008dataset} is central to various long standing research directions, such as domain adaptation~\citep{ben2006analysis}, domain generalization~\citep{blanchard2021domain}, meta-learning~\citep{hochreiter2001learning,hospedales2021meta}, and active learning~\citep{settles2009active}.
To counteract these shifts, a common goal is to adapt a model trained on a source distribution to a shifted target domain.
For engineering tasks, where rapid adaptation is essential and target domain distributions are unavailable a priori, \acf{TTA} is a particularly suitable approach, as it adapts models during inference without access to source data and with minimal computational overhead~\citep{liang2020we, sun2020test, wang2020tent}.
\ac{TTA} has proven effective in many domains, including medical imaging, object detection, and segmentation.

However, while many works treat classification tasks~\citep{wang2020tent, liang2020we}, comparably little research can be found for high-dimensional regression settings~\citep{liang_comprehensive_2025}.
One interesting recent method tackling this gap is \ac{SSA}~\citep{adachi_ssa_2025}, capable of handling both classification and one-dimensional regression tasks.
Further methods that are similar to the high-dimensional simulation tackle \ac{TTA} in vision settings, for example depth-estimation~\citep{liu2023ttadepthpred}, super-resolution~\citep{park2020superresolutionmetalearning, deng2023ttasuperresolution} and image dehazing~\cite{liu2022ttaimagedehazing}.
Unfortunately, first, all methods mentioned above are tailored to the image domain, relying on inductive biases of regular grids and visual similarities between inputs and outputs.
This does not translate to the unstructured non-euclidean domains of engineering simulations, where the compact inputs like scalar simulation parameters and geometries can generate complex solution manifolds.
Second, the methods mentioned above typically operate on problems of $\mathcal{O}(10^5)$ degrees of freedom (e.g. images up to $512\times512$), whereas the numerical simulation grids we consider can reach up to $\mathcal{O}(10^6)$ and further increase orders of magnitudes in industry~\citep{ferziger2019computational}.
As a consequence, classical TTA methods often cannot overcome the severe instabilities in our considered problem setting.

To address this, we introduce a (to the best of our knowledge first) \ac{TTA} framework explicitly targeting neural surrogates for high-dimensional engineering tasks under distribution shifts, covering simulation (regression) and design optimization (generation) problems.
At the core of our approach lies the use of maximally informative source statistics to stabilize the adaptation process, which we approach by D-optimal~\citep{atkinson1992optimum} sample selection.
This approach allows us to compress the source manifold into a small set of source statistics for realizing three core properties for robust \ac{TTA}:
\begin{enumerate*}[label=(\roman*)]
  \item feature alignment,
  \item preservation of source domain knowledge, and
  \item unsupervised tuning of adaptation hyperparameters.
\end{enumerate*}

To achieve (i), we extend one-dimensional regression approaches for covariance adaptation~\citep{adachi_ssa_2025} to our multidimensional setting.
This places our work within the broader category of domain-invariant representation learning algorithms~\citep{ben2006analysis,ganin2015dann,cmd,johansson2019support}.
To ensure (ii), we regularize adaptation using source error induced by our D-optimal statistics, effectively constraining updates to remain close to the source solution.
Finally, tackling (iii) is crucial as the optimal choice of adaptation hyperparameters has been acknowledged as a big bottleneck in the area of \ac{UDA} and related communities~\citep{musgrave2021realitycheck,miller2021accuracy_on_the_line,baek2023agreementonthelinepredictingperformanceneural,setinek2025simshift}.
We therefore integrate Importance Weighted Validation (IWV) using estimated density ratios as an unsupervised model selection strategy, enabling automated, in-the-loop model selection in order to optimize performance while ensuring stability at the same time.

We summarize our contributions as follows:
\begin{itemize}
    \item \textit{Problem:} We are (to the best of our knowledge) the first one applying TTA to high-dimensional simulation regression.
    \item \textit{Method:} We propose a novel adaptation framework that relies on D-optimal source statistics and stabilizes three main components: feature alignment, source knowledge preservation, and parameter tuning.
    \item \textit{Performance:} We demonstrate in Table \ref{tab:comparison_simshift} and \ref{tab:comparison_engibench} that our approach reliably outperforms standard \ac{TTA} methods on diverse engineering adaptation benchmarks, SIMSHIFT for high-dimensional regression and EngiBench for design generation.
\end{itemize}

\begin{figure*}[t]
    \centering
    \includegraphics[width=0.85\textwidth,trim=0 50 30 20,clip]{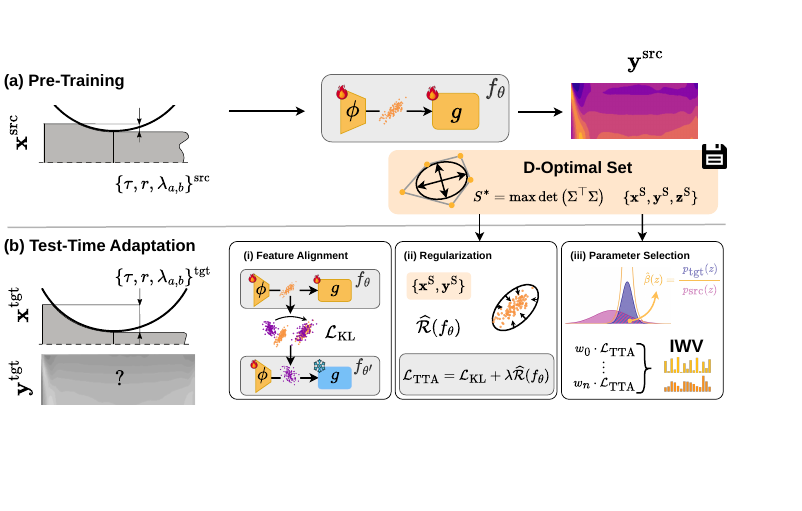}
    \vspace{0.2em}
    \caption{
    Our method applied to \textit{hot rolling} task from~\citep{setinek2025simshift}. 
    (a) Pre-training on the source domain with fixed input parameters, thickness ($\tau$), post-rolling reduction ($r$), and temperature coefficients ($\lambda_a$, $\lambda_b$). The \textit{representation learner} $\phi$ and the \textit{predictor} $g$ are optimized and maximally informative (D-optimal) statistics are computed. 
    (b) Test-time adaptation of $\phi$ without source data using D-optimal statistics for realizing three TTA pillars: adaptation (KL-based feature alignment), source knowledge preservation (statistics-based regularization) and parameter tuning (importance weighted validation).
    }
    \label{fig:adaptation}
\end{figure*}

\section{Related Work}
\label{sec:relwork}
\textbf{Neural surrogates} have emerged as a widely used approach to accelerate traditional numerical simulation methods by providing fast approximations of the solutions.
In general, surrogate models are trained on the solutions from numerical solvers, paired with the corresponding initial conditions and configurations under which they were generated, e.g., \cite{setinek2025simshift, bonnet2022airfrans, toshev2023lagrangebench, toshev2024jaxsph}.
A particularly prominent line of work within neural surrogate modeling for \acp{PDE} is operator learning \citep{kovachki2021neuraloperator, li2020gkn, lu2021deeponet, alkin2024upt, wu2024Transolver}. Such models aim to directly approximate the solution operator that maps initial functions (conditions and input terms) to output functions.

\textbf{\acf{TTA}} refers to the emerging machine learning technique of adapting a pre-trained model to unlabeled target data, directly at inference time and prior to generating predictions.
For this reason, \ac{TTA} has recently attracted increasing attention as it can offer (nearly) free performance gains \citep{liang_comprehensive_2025}.
While the majority of existing \ac{TTA} methods have been developed for low-dimensional classification tasks \citep{liang2021reallyneedaccesssource, yang2021exploitingintrinsicneighborhoodstructure}, employing methodologies such as entropy minimization \citep{wang2020tent, zhou2021bayesian, niu2022efficient, zhang2022memo, zhao2023delta} and feature alignment \citep{ishii2021source, kojima2022robustifying, eastwood2021source, adachi2023covariance, jung2023cafa}, recent works have begun to extend these ideas to image segmentation
\citep{valanarasu2023on_the_fly_tta, HE2021102136, Karani_2021}.
Research in \ac{TTA} tackling regression problems is much sparser.
\acf{SSA} \citep{adachi_ssa_2025} moves into this direction, showing positive performance in the one-dimensional cases. Additionally, there are specialized methods designed for image regression tasks such as depth-estimation~\citep{liu2023ttadepthpred}, super-resolution~\citep{park2020superresolutionmetalearning, deng2023ttasuperresolution}, or image dehazing~\cite{liu2022ttaimagedehazing}.
Finally, \ac{TTA} should not be confused with Test-Time Training (TTT), often used in time series literature \citep{pmlr-v119-sun20b, wang2020tent, sun2025learninglearntesttime, sun2020testtimetrainingselfsupervisiongeneralization}. While both solve the same problem, TTT typically refers to methods that employ time-series specific techniques, for example, updating hidden states during sequential inference.

\textbf{Covariance alignment} of latent feature distributions is common practice in \ac{UDA} and \ac{TTA}~\citep{coral,coral_og,Li2016RevisitingBN,wang2020tent}.
Even though this can be extended to higher-order moments~\citep{cmd,chen2019hommhigherordermomentmatching}, second-order alignment often already achieves stable performance across datasets.

\textbf{Domain generalization, meta-learning, and active learning} represent alternative strategies that can be used to improve model robustness and generalization under distribution shifts. 
Domain generalization \citep{pmlr-v28-muandet13, li2017learninggeneralizemetalearningdomain,holzleitner2024domain}
and \ac{UDA} \citep{coral, mmd, cmd, ganin2015dann} can be effective in some scenarios, however their reliance on specific training, model selection and diverse training distributions limits their applicability.
Meta-learning methods \citep{hochreiter2001learning,finn2017maml} and active learning \citep{lewis1994uncertainty, al4pde} are similarly motivated, but generally assume access to ground-truth information in the shifted domain.
In our setting, all these approaches face a significant practical limitation: none of them can quickly adapt a pre-trained model leveraging unlabeled data at test-time, as they all rely on a priori knowledge and training. This motivates our exploration of \ac{TTA} as a more suitable solution.

\section{Problem}
\label{sec:problem}

Following~\citep{xiao2024beyond,liang_comprehensive_2025}, we assume access to a regressor $f_\theta : \mathcal{X} \to \mathbb{R}^d$ pre-trained on \textit{source} samples $(\mathbf{x}_i, \mathbf{y}_i)_{i=1}^{N^{\mathrm{src}}}\in \mathcal{X}\times \mathbb{R}^d$ drawn from a source distribution $P^{\mathrm{src}}$, e.g., $f_\theta=g\circ \phi$ in \cref{fig:adaptation}.
We also assume access to some ground truth matrix-valued source statistics.

The goal is, for any new \textit{unlabeled} sample $(\mathbf{x}_i^{\mathrm{tgt}})_{i=1}^{N^\mathrm{tgt}}$ drawn from the input marginal of a \textit{target} distribution $P^{\mathrm{tgt}} \neq P^{\mathrm{src}}$, to find $\theta$ which minimizes the empirical target risk
\begin{equation}
\widehat{\mathcal{R}}_\mathrm{tgt}(f_\theta)=\frac{1}{N^{\mathrm{tgt}}} \sum_{i=1}^{N^{\mathrm{tgt}}} \left\lVert f_\theta(\mathbf{x}_i^{\mathrm{tgt}}) - \mathbf{y}_i^{\mathrm{tgt}}\right\rVert_2^2.
\label{eq:target_risk}
\end{equation}
Note that we have no access to any target labels $(\mathbf{y}_i^{\mathrm{tgt}})_{i=1}^{N^{\mathrm{tgt}}}$ and the target risk in Eq.~\eqref{eq:target_risk} cannot be directly evaluated.

\textbf{\ac{TTA} in simulation.}
We emphasize that our \ac{TTA} setting differs from the usual, computer vision-oriented problems. In particular, simulation surrogates are more challenging, as the output dimension $d$ of $f_\theta : \mathcal{X}\!\to\!\mathbb{R}^d$ can reach $O(10^6)$ in our regime.
This is typical for neural surrogates in simulation, but less common in the vision domain. 
Moreover, adaptation relies on small unlabeled target batches $\{\mathbf{x}_i^{\mathrm{tgt}}\}_{i=1}^{N_{\mathrm{tgt}}}$ with $N_{\mathrm{tgt}}\!\ll\! d$.
Finally, vision tasks often present structured, visually aligned inputs and outputs. Conversely, simulation data usually lacks geometric correspondence between $\mathbf{x}_i$ and $\mathbf{y}_i$, as $\mathbf{x}_i$ is often just coordinates \citep{lu2021deeponet, kovachki2021neuraloperator}.
This, together with the high dimensionality render standard \ac{TTA} methods ill-conditioned for neural surrogates, and necessitates explicit methodological mechanisms to stabilize the adaptation process.

\section{Method}
\label{Method}

\subsection{Maximally informative statistics}
In high-dimensional settings, naive statistics (e.g., global means) are insufficient to support reliable \ac{TTA}, as their estimation becomes ill-conditioned in the presence of low-information or spurious feature directions. As a result, usual adaptation objectives become sensitive to noise and irrelevant components. We approach this by selecting a subset of latent representations that preserves the most informative structure of the source model \cite{zhang2023improving}.

\textbf{Data generating assumption.} We follow the common assumption~\citep{subspace2014, coral, adachi_ssa_2025} that $\mathbf{z}=\phi(\mathbf{x})$ is normally distributed for each domain, such that the feature distribution is fully characterized by its first- and second-order moments (mean, covariance). 

\textbf{D-optimal latent statistics.}
Under this assumption, we focus on second-order latent statistics and select source samples that maximize the information retained in the latent space via D-optimality~\citep{atkinson1992optimum}.
Originating in experimental design, D-optimality identifies a subset of $m$ samples whose (latent) representations span the most informative subspace of the original (feature) space.
In our setting, letting $\mathbf{Z}_S \in \mathbb{R}^{m \times d}$ denote the matrix of latent features $\mathbf{z}=\phi(\mathbf{x})$ for a subset $S \subset \{1,\dots,N\}$, we select
\[
S^\star = \arg\max_{|S|=m} \det\!\left(\mathbf{Z}_S^\top \mathbf{Z}_S\right),
\]
which equivalently maximizes the volume of the linear subspace spanned by the selected vectors.
Note that when the latent features are centered, $\mathbf{Z}_S^\top \mathbf{Z}_S$ is proportional to the empirical covariance matrix of the selected samples. This means that maximizing the determinant of $\mathbf{Z}_S^\top \mathbf{Z}_S$ corresponds to maximizing the generalized variance of the retained latent representation.


For tractability, we follow the approach of approximating the D-optimal criterion using QR pivoting on whitened principal components \citep{golub2013matrix}. See \cref{alg:doptimal} for the pseudocode of our Quasi D-optimal selection criteria.

\begin{algorithm}
\caption{Quasi D-optimal spanning set selection via PCA and QR pivoting. \label{alg:doptimal}}
\begin{algorithmic}[1]
\REQUIRE Inputs $\mathbf{x}^{src}$, eigendecomposition $(\boldsymbol{\lambda},\mathbf{V})$ of $\phi(\mathbf{x}^{src})$, variance threshold $\tau$, number of quasi D-optimal designs $m$ 
\ENSURE Selected source dataset indices $S \subseteq \{1, \dots, N\}$
\STATE $\mathbf{Z} \leftarrow \phi(\mathbf{x}^{src}) $
\STATE $\mathbf{Z} \leftarrow \mathbf{Z} - \text{mean}(\mathbf{Z})$
\STATE $r \leftarrow \text{select\_components}(\boldsymbol{\lambda}, \tau)$   \COMMENT{keep $\tau\%$ variance}
\STATE $\mathbf{Y} \leftarrow \mathbf{Z} \mathbf{V}_{:,1:r} \boldsymbol{\Lambda}_r^{-1/2}$
\STATE $\mathbf{Q}, \mathbf{R}, \text{piv} \leftarrow \text{QR}(\mathbf{Y}^T)$
\STATE $S \leftarrow \text{piv}_{1:m}$
\STATE \textbf{return} $S$
\end{algorithmic}
\end{algorithm}

\subsection{SATTS}
\label{subsec:featurealign}
We term our approach Stable Adaptation at Test-Time for Simulation (SATTS).
SATTS uses D-optimal statistics in a unified framework for stabilizing TTA of high-dimensional simulation regressors.
More precisely, D-optimal statistics are used at three key \ac{TTA} components: feature alignment, source knowledge preservation, and parameter tuning.

\textbf{Feature alignment} is a common approach in \ac{TTA} to reduce the dissimilarity between source and target distributions (Section~\ref{sec:relwork}).
We design our regressor as $f = g \circ \phi$, where a \textit{representation learner} $\phi$ maps inputs to latent features $\mathbf{z} \in \mathbb{R}^C$, and a \textit{predictor} $g$ maps these features to outputs, see Figure~\ref{fig:adaptation}.
During test-time adaptation, only the representation learner $\phi$ is updated, while the predictor $g$ remains fixed.

In our high-dimensional regression setting, predictions are given by
\begin{align*}
g(\mathbf{z}) = \mathbf{W}\mathbf{z} + \mathbf{b},
\end{align*}
where $\mathbf{W} \in \mathbb{R}^{K \times C}$ maps latent features to $K$ output dimensions.

Following a recent TTA method for regression, Significant Subspace Alignment (SSA)~\citep{adachi_ssa_2025}, latent directions that strongly influence the prediction are the right candidates for alignment.
In SSA, feature importance is defined for one-dimensional regression by selecting a subset of principal directions based on $|w^\top v_k|$, resulting in a hard truncation to a manually chosen significant subspace.

We generalize this idea to high-dimensional regression by assigning a \emph{positive importance weight} to every principal direction,
\begin{equation}
\label{eqs:alpha}
    \alpha_k = 1 + \left\lVert \mathbf{W}\mathbf{v}_k^{\mathrm{src}} \right\rVert_2,
\end{equation}
where $\mathbf{v}_k^{\mathrm{src}} \in \mathbb{R}^C$ denotes the $k$-th source principal component.
Notably, when $K=1$ and only a subset of directions is retained, this formulation reduces to SSA.

At deployment, target features $\mathbf{z}^{\mathrm{tgt}}=\phi(\mathbf{x}^{\mathrm{tgt}})$ are centered using the source mean $\mu^{\mathrm{src}}$, projected onto the source principal components $\mathbf{V}^{\mathrm{src}}=[\mathbf{v}_1^{\mathrm{src}},\dots,\mathbf{v}_K^{\mathrm{src}}]$, and reweighted by $\boldsymbol{\alpha}$:
\begin{equation}
\label{eqs:features}
    \tilde{\mathbf{z}}^{\mathrm{tgt}}
    =
    \big(\mathbf{z}^{\mathrm{tgt}} - \mu^{\mathrm{src}}\big)\mathbf{V}^{\mathrm{src}} \odot \boldsymbol{\alpha}.
\end{equation}

Finally, we align source and target feature distributions under the data generating assumption by minimizing a channel-wise symmetric empirical KL-divergence between the projected target statistics $(\tilde{\mu}_k^{\mathrm{tgt}}, \tilde{\sigma}_k^{\mathrm{tgt}2})$ and the (D-optimally stabilized) source statistics $(0, \lambda_k^{\mathrm{src}})$:
\begin{equation}
\label{eqs:kl_loss}
\mathcal{L}_{\mathrm{KL}}
=
\frac{1}{2}
\sum_{k=1}^{K}
\left(
\frac{(\tilde{\mu}_k^{\mathrm{tgt}})^2 + \lambda_k^{\mathrm{src}}}{\tilde{\sigma}_k^{\mathrm{tgt}2}}
+
\frac{(\tilde{\mu}_k^{\mathrm{tgt}})^2 + \tilde{\sigma}_k^{\mathrm{tgt}2}}{\lambda_k^{\mathrm{src}}}
-2
\right).
\end{equation}
In contrast to SSA, which relies on hard selection of a manually chosen significant subspace, our formulation assigns strictly positive weights to all principal directions, resulting in a soft and dense reweighting of the latent space.
Crucially, all source statistics required for the weighting and subsequent alignment are computed exclusively from the D-optimal subset, ensuring that the induced covariance structure remains maximally informative and well-conditioned.
This is essential for stabilizing feature alignment in very high-dimensional regression (see Section~\ref{sec:experiments}).

\textbf{Source knowledge preservation}
is realized by regularization on the subsampled source statistics:
\begin{equation} \label{eq:combined_loss}
    \mathcal{L}_\mathrm{TTA} = \mathcal{L}_{\mathrm{KL}} + \lambda \widehat{\mathcal{R}}_\mathrm{src}(f_\theta)
\end{equation}
with $\hat{\mathcal{R}}_{\mathrm{src}}$ denoting the empirical source risk estimated on the D-optimal samples and $\lambda>0$ being a regularization parameter.
This ensures that the feature alignment updates driven by $\mathcal{L}_{\mathrm{KL}}$ do not deviate significantly from the known solution.

\begin{table*}[t]
\small
\centering
\caption{Comparison of current baselines with TTA methods for all simulation datasets. Results are averaged across 20 \ac{TTA} runs, over a pretrained model with standard deviation reported. Reported RMSE is normalized over all fields.}
\label{tab:comparison_simshift}

\begin{subtable}{0.44\linewidth}
\centering
\caption{Rolling}
\resizebox{\linewidth}{!}{%
\begin{tabular}{lccc}
\toprule
Model & RMSE ($\downarrow$) & MAE ($\downarrow$) & $R^2$ ($\uparrow$) \\
\midrule
Source & $0.561_{\pm 0.001}$ & $0.484_{\pm 0.001}$ & $0.781_{\pm 0.001}$ \\
\midrule
Tent & $1.825_{\pm 0.002}$ & $1.553_{\pm 0.002}$ & $-0.371_{\pm 0.004}$ \\
SSA & $0.566_{\pm 0.020}$ & $0.481_{\pm 0.018}$ & $0.811_{\pm 0.014}$ \\
\ourmethod & $\mathbf{0.545_{\pm 0.019}}$ & $\mathbf{0.466_{\pm 0.018}}$ & $\mathbf{0.831_{\pm 0.012}}$ \\
\midrule
Oracle & $0.529_{\pm0.013}$ & $0.453_{\pm0.012}$ & $0.832_{\pm0.011}$\\
\bottomrule
\end{tabular}}
\end{subtable}
\hspace{3.5em}
\begin{subtable}{0.44\linewidth}
\centering
\caption{Motor}
\resizebox{\linewidth}{!}{%
\begin{tabular}{lccc}
\toprule
Model & RMSE ($\downarrow$) & MAE ($\downarrow$) & $R^2$ ($\uparrow$) \\
\midrule
Source & $0.109_{\pm 0.001}$ & $0.058_{\pm 0.001}$ & $0.989_{\pm 0.001}$ \\
\midrule
Tent & $1.132_{\pm 0.032}$ & $0.753_{\pm 0.026}$ & $-0.152_{\pm 0.065}$ \\
SSA & $0.336_{\pm 0.001}$ & $0.172_{\pm 0.006}$ & $0.881_{\pm 0.008}$ \\
\ourmethod & $\mathbf{0.109_{\pm 0.003}}$ & $\mathbf{0.058_{\pm 0.001}}$ & $\mathbf{0.989_{\pm 0.000}}$ \\
\midrule
Oracle & $0.108_{\pm 0.001}$ & $0.058_{\pm 0.001}$ & $0.989_{\pm 0.001}$ \\
\bottomrule
\end{tabular}}
\end{subtable}
\\
\vspace{2em}
\small
\begin{subtable}{0.44\linewidth}
\centering
\caption{Forming}
\resizebox{\linewidth}{!}{%
\begin{tabular}{lccc}
\toprule
Model & RMSE ($\downarrow$) & MAE ($\downarrow$) & $R^2$ ($\uparrow$) \\
\midrule
Source & $0.161_{\pm 0.001}$ & $0.066_{\pm 0.001}$ & $0.979_{\pm 0.001}$ \\
\midrule
Tent & $1.251_{\pm 0.001}$ & $0.639_{\pm 0.001}$ & $-0.081_{\pm 0.001}$ \\
SSA & $0.215_{\pm 0.005}$ & $0.098_{\pm 0.003}$ & $0.965_{\pm 0.002}$ \\
\ourmethod & $\mathbf{0.157_{\pm 0.001}}$ & $\mathbf{0.066_{\pm 0.001}}$ & $\mathbf{0.980_{\pm 0.001}}$ \\
\midrule
Oracle & $0.156_{\pm0.004}$ & $0.067_{\pm0.002}$ & $0.980_{\pm0.001}$\\ 
\bottomrule
\end{tabular}}
\end{subtable}
\hspace{3.5em}
\begin{subtable}{0.44\linewidth}
\centering
\caption{Heatsink}
\resizebox{\linewidth}{!}{%
\begin{tabular}{lccc}
\toprule
Model & RMSE ($\downarrow$) & MAE ($\downarrow$) & $R^2$ ($\uparrow$) \\
\midrule
Source & $0.747_{\pm 0.001}$ & $0.565_{\pm 0.001}$ & $0.237_{\pm 0.001}$ \\
\midrule
Tent & $0.876_{\pm 0.001}$ & $0.694_{\pm 0.0}$ & $-0.203_{\pm 0.007}$ \\
SSA      & $0.746_{\pm 0.001}$ & $0.552_{\pm 0.001}$ & $0.227_{\pm 0.001}$ \\
\ourmethod & $\mathbf{0.738_{\pm 0.004}}$ & $\mathbf{0.545_{\pm 0.003}}$ & $\mathbf{0.244_{\pm 0.007}}$ \\
\midrule
Oracle & $0.732_{\pm 0.035}$ & $0.541_{\pm 0.03}$ & $0.265_{\pm 0.065}$\\ 
\bottomrule
\end{tabular}}
\end{subtable}
\end{table*}

\textbf{Parameter tuning}
We integrate Importance Weighted Validation (IWV)~\citep{shimodaira2000improving} using D-optimal samples to tune the test-time adaptation learning rate.
Since the target risk in Eq.~\eqref{eq:target_risk} cannot be computed directly without access to target labels, we estimate it via an importance-weighted source risk under the covariate shift assumption $p_{\mathrm{src}}(\mathbf{y} \mid \mathbf{x}) = p_{\mathrm{tgt}}(\mathbf{y} \mid \mathbf{x})$:
\begin{equation}
\label{eq:iwr}
\hat{\mathcal{R}}_{\mathrm{tgt}}(f_\theta)
\;\approx\;
\frac{1}{m}
\sum_{i=1}^{m}
\hat{\beta}(\mathbf{z}_i)\,
\lVert f_\theta(\mathbf{x}_i^S) - \mathbf{y}_i^S \rVert_2^2 ,
\end{equation}
where $\{(\mathbf{x}_i^S,\mathbf{y}_i^S)\}_{i=1}^{m}$ denotes the set of D-optimal source samples and
$\hat{\beta}(\mathbf{z}) = p_{\mathrm{tgt}}(\mathbf{z}) / p_{\mathrm{src}}(\mathbf{z})$ is the density ratio estimated in latent space $\mathbf{z}=\phi(\mathbf{x})$ under the Gaussian data generating assumption.
Using the estimate in Eq.~\eqref{eq:iwr}, we perform model selection via line search over the TTA learning rate, evaluating performance after each adaptation step and stopping once further updates no longer improve the objective.


\section{Experiments}
\label{sec:experiments}
This section presents an empirical analysis of Stable Adaptation at Test-Time for Simulation. We introduce the benchmarks used for evaluation and demonstrate the performance of our method in complex industrial use cases, where neural surrogates are employed to approximate costly numerical simulations or to directly generate candidate designs. In all experiments, the parameters of the quasi D-optimal algorithm (see \cref{alg:doptimal}) are fixed with $m = 8$ indices and a threshold of $\tau = 0.95\%$.

\subsection{Datasets}
Our evaluation is conducted on two simulation benchmarks, SIMSHIFT \citep{setinek2025simshift} and EngiBench \citep{felten2025engibenchframeworkdatadrivenengineering}.
SIMSHIFT is designed to evaluate how surrogate models adapt to distribution shifts on real-world industrial simulation tasks, while EngiBench is a collection of design optimization datasets, optimizers, and simulators to evaluate designs.
In both benchmarks, the inputs $\bf x$ represent parameters like geometry, material properties, desired or operating conditions.
The ``labels'' $\bf y$ correspond to high-dimensional fields such as stresses or deformation for SIMSHIFT, and material density of the generated design for EngiBench.

The target distributions in both cases are generated from unseen parameter configurations, and the goal is to predict the corresponding fields outside the training regime. While SIMSHIFT formulates the problem as a regression task with neural operators \citep{kovachki2021neuraloperator}, EngiBench treats it as an inverse problem solved by generative models.
The diversity in task formulation and training paradigm across the two benchmarks highlights the model-agnostic nature of our method.

\subsection{Neural Surrogates for Simulation: SIMSHIFT}\label{subsec:SimshiftExperiments}

We first analyze adaptation behavior on the SIMSHIFT benchmark \citep{setinek2025simshift}. SIMSHIFT spans four distinct industrial simulation settings: \textit{hot rolling}, \textit{sheet metal forming}, \textit{electric motor}, and \textit{heatsink design}. All datasets have explicit source and target domain splits, dependent on the parameters such as initial conditions, material or geometry specifications that were used to generate the samples.
Shifts happen in parametric space, as opposed to unstructured variations occurring in images.
We perform all our experiments using the medium difficulty domain shift setup for all datasets. 
For a detailed description of the datasets, their creation, and the defined distribution shifts, we refer the reader to the SIMSHIFT publication \citep{setinek2025simshift}.

\begin{figure}[b!]
    \centering
    \includegraphics[width=\linewidth]{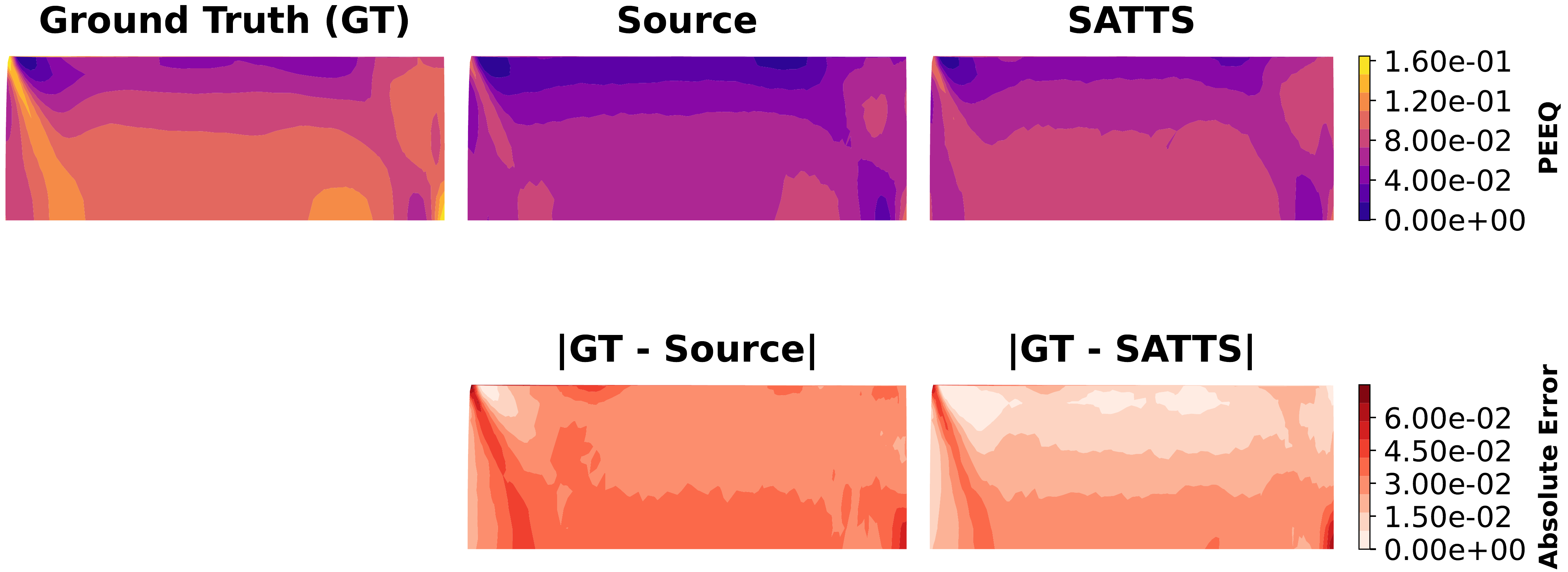}
    \vspace{0.2em}
    \caption{Comparison of Equivalent Plastic Strain (PEEQ) predictions on a hot rolling sample. Displaying the Ground Truth (GT), the unadapted \textit{Source} model, and the \ourmethod~results in the top row, with the absolute residuals, $|\text{GT} - \text{Source}|$ and $|\text{GT} - \text{\ourmethod}|$ in the bottom row.}
    \label{fig:sample_rolling}
\end{figure}

\cref{tab:comparison_simshift} summarizes the results across all datasets, comparing our method with \ac{SSA} and Tent as established \ac{TTA} baselines, as well as the unadapted source model (\textit{Source}) and the target-optimal selection (\textit{Oracle}). Implementation details are provided in \cref{apd:setup}.

Across all settings, \ourmethod~consistently outperforms \ac{SSA} and yields the strongest performance among all adaptation methods, establishing a new baseline for test-time adaptation in neural surrogate regression. While the absolute gains over the source model are modest in some regimes, they are achieved without sacrificing stability. In contrast, both \ac{SSA} and, more prominently, Tent frequently degrade performance relative to the pre-trained model, indicating a lack of robustness to these high-dimensional distribution shifts.

Additionally, visual analysis of the Equivalent Plastic Strain (PEEQ) for the \textit{hot rolling} dataset in Figure \ref{fig:sample_rolling} reveals that \ourmethod, successfully corrects systematic under-predictions in the deformation zones. This indicates that the adapted model ensures better physical consistency with the ground truth.

\begin{figure}[!b]
    \centering
    \includegraphics[width=\linewidth]{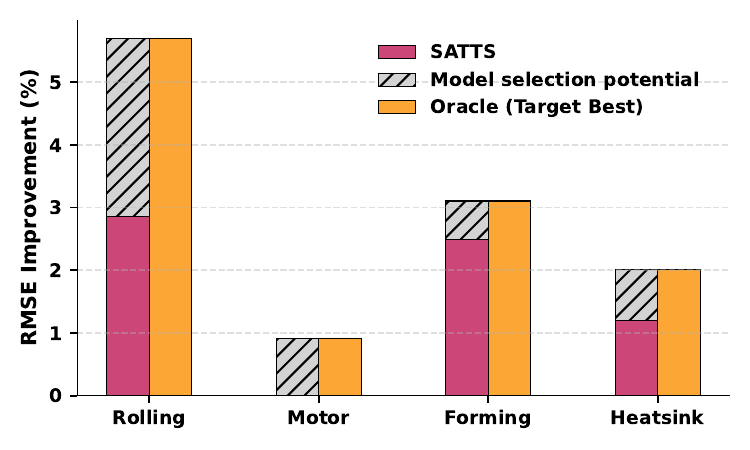}
    \caption{Relative performance improvements of \ourmethod~and the Oracle (lower bound for model selection) compared to the Source model, measured by RMSE.}
    \vspace{0.2em}
    \label{fig:rel_improvement_tta}
\end{figure}

Finally, the comparison to the \textit{Oracle} highlights that \ourmethod~ substantially reduces the performance gap to the target-optimal solution, though it does not fully close it, suggesting that further gains may be possible with stronger unsupervised model selection strategies.

To improve the interpretability of our results, \cref{fig:rel_improvement_tta} displays the relative performance improvements of our method compared to the \textit{``unregularized''} pre-trained (\textit{Source}) model for all dataset and highlights the potential of perfect model selection.
Additionally, we quantify the discrepancy between the source and target domains directly in the output space.
Namely, the potential transfer error is upper bounded by the $\mathcal{H}$-divergence which itself is upper bounded by the Proxy $\mathcal{A}$-distance (PAD) (for details see \citet{bouvier2020robust_da}, \citet{johansson2019support} and \citet{zellinger2021balancing}).
We estimate PAD by training a domain classifier directly on the simulation outputs and converting its test error into a distance estimate.
The resulting PAD values for all simulation datasets are provided in \cref{tab:pad_values}.
Comparing these distances with the performance gains shown in \cref{fig:rel_improvement_tta}, we observe datasets with larger PAD values all exhibit performance improvements from adaptation.
In contrast, the \textit{motor} dataset shows the smallest PAD, indicating a comparatively weak output-space shift, which is consistent with no performance improvements observed in this case.

\begin{table}[ht]
\renewcommand{\arraystretch}{1.0}
\caption{PAD values for the simulation datasets.}
\label{tab:pad_values}
\centering
\resizebox{0.4\textwidth}{!}{
\begin{tabular}{lcccc}
\toprule
\textbf{Dataset} & \textbf{Rolling} & \textbf{Forming} & \textbf{Motor} & \textbf{Heatsink} \\
\midrule
\textbf{PAD} & 1.765 & 0.929 & 0.314 & 1.767 \\
\bottomrule
\end{tabular}
}
\end{table}

\subsection{Generative Design Optimization: EngiBench}
\label{subsec:EngiExperiments}

\begin{table*}[t!]
\small
\centering
\caption{
Comparison of current baselines with TTA methods for design optimization datasets. Results are averaged across 20 \ac{TTA} runs, over one model with standard deviation reported.}
\label{tab:comparison_engibench}
\begin{subtable}{0.45\linewidth}
\centering
\caption{Beams2D}
\begin{tabular}{lccc}
\toprule
Model & COMP ($\downarrow$) & MAE ($\downarrow$) & MMD ($\downarrow$) \\
\midrule
Source & $123.7_{\pm 17.854}$ & $\mathbf{0.026_{\pm 0.004}}$ & $\mathbf{0.052_{\pm 0.002}}$ \\ 
\midrule
SSA & $119.4_{\pm 4.586}$ & $0.040_{\pm 0.005}$ & $0.062_{\pm 0.003}$ \\ 
\ourmethod   & $\mathbf{118.8_{\pm 12.409}}$ & $0.027_{\pm 0.004}$ & $0.053_{\pm 0.002}$ \\ 
\midrule
Oracle & $113.8_{\pm 1.267}$ & $0.026_{\pm 0.003}$ & $0.038_{\pm 0.001}$ \\ 
\bottomrule
\end{tabular}
\end{subtable}
%
\hspace{3.0em}
\begin{subtable}{0.45\linewidth}
\centering
\caption{HeatConduction2D}
\begin{tabular}{lccc}
\toprule
Model & COMP ($10^{-3}$) & MAE ($\downarrow$) & MMD ($\downarrow$) \\
\midrule
Source & $0.577_{\pm 0.561}$ & $0.336_{\pm 0.057}$ & $0.095_{\pm 0.000}$ \\
\midrule
SSA & $0.712_{\pm 0.615}$ & $0.349_{\pm 0.057}$ & $0.095_{\pm 0.000}$ \\
\ourmethod  & $\mathbf{0.537_{\pm 0.491}}$ & $\mathbf{0.334_{\pm 0.015}}$ & $0.095_{\pm 0.000}$ \\
\midrule
Oracle & $0.509_{\pm 0.416}$ & $0.329_{\pm 0.052}$ & $0.095_{\pm 0.000}$ \\
\bottomrule
\end{tabular}
\end{subtable}
\end{table*}

We evaluate on two EngiBench design optimization tasks: \emph{structural beam bending} and \emph{2D heat conduction}.
By default, these datasets do not include predefined source and target domains.
We therefore define them following the approach in \cite{setinek2025simshift}: we train models on the full datasets and subsequently analyze the t-SNE visualizations of the latent feature spaces as the input conditions are varied.
Datasets are then partitioned into source and target domains based on the parameters that dominate the latent space variation.
A detailed analysis of this procedure and corresponding visualizations can be found in \cref{app:engibench_splits}.

We report \ac{MAE}, the \ac{MMD}, and Compliance (COMP), a dataset specific objective value calculated with a \ac{FEM} solver.
For Beams2D, compliance is the inverse of stiffness whereas for HeatConduction2D it is the thermal compliance coefficient.
All performance metrics reflect only feasible design solutions, as we exclude structural failure cases for analysis.

In \cref{tab:comparison_engibench}, we compare our method against the \textit{``unregularized''} pre-trained model (\textit{Source}), \ac{SSA} and the target-optimal selection (\textit{Oracle}).
Across both tasks, our approach typically matches or reduces errors relative to the unregularized model.
Compared to our method, \ac{SSA} shows unstable behavior on certain metrics, sometimes even deteriorating performance substantially.
Such behavior is highly undesirable in \ac{TTA} deployments and underlines the strong suit of our approach: its stability.

\begin{figure}[b!]
    \centering
    \includegraphics[width=\linewidth]{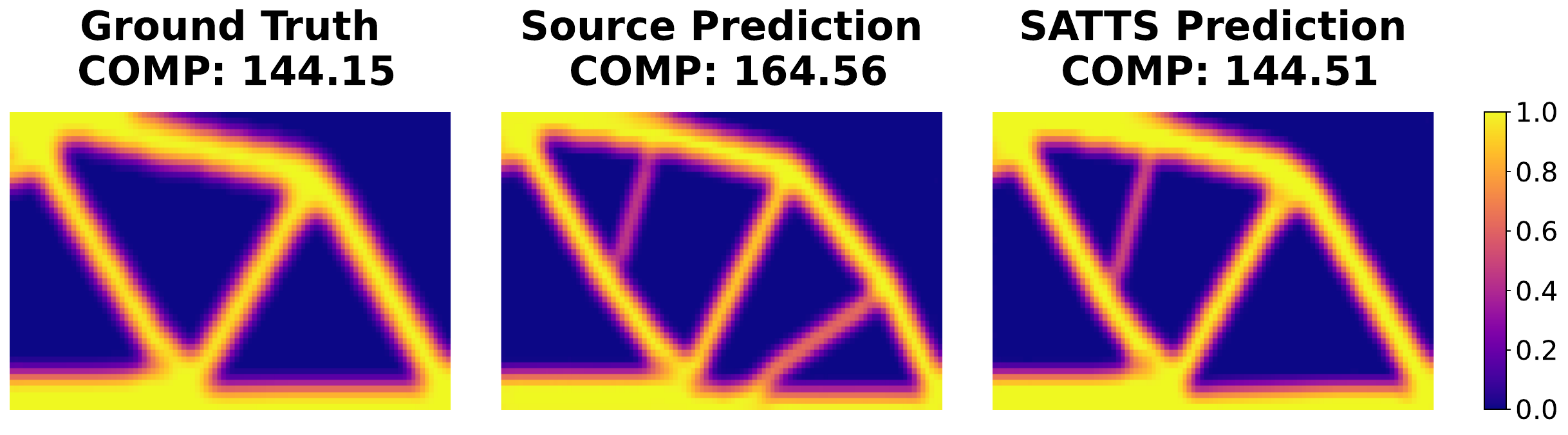}
    \caption{Comparison of 2D beam topology results based on ground truth, source prediction, and \ourmethod. The heatmaps illustrate material density $\rho \in [0, 1]$. \ourmethod~shows stronger alignment with the original design, resulting in more robust design outcomes.}
    \label{fig:sample_beams}
\end{figure}

Figure \ref{fig:sample_beams} undermines the numerical results, as \ourmethod~produces superior 2D beam design outputs when compared to the pre-trained (\textit{Source}) model. 


\subsection{Ablations}
\textbf{Component Analysis} To assess the contribution of D-optimal source selection and importance weighting, we perform an incremental ablation study. Starting from the existing alignment strategy SSA, we isolate the effect of D-optimal source importance weighting from source selection. For both the SSA baseline and the source importance weighting, we use the original parameter values ($lr = 0.01$) from \citet{adachi_ssa_2025}. Results on the SIMSHIFT benchmark in \cref{tab:methodablation} show that each incremental addition improves performance over the previous configuration. 

\begin{table}[h]
\small
\centering
\caption{RMSE scores of \ourmethod~with and without importance weighting and model selection. The baseline and IWV results were evaluated with $lr = 0.01$. Best scores are bolded.}
\begin{subtable}{\linewidth}
\label{tab:methodablation}
\centering
\resizebox{\linewidth}{!}{%
\begin{tabular}{cccccc}
\toprule
$\hat{\mathcal{R}}_\mathrm{src}(f_\theta)$ & 
IWV & Rolling &  Motor & Forming \\ \midrule
         &  &  $0.566_{\pm0.020}$  &  $0.336_{\pm0.000}$ & $0.215_{\pm0.005}$ \\
\checkmark &      &$0.550_{\pm0.020}$ &  $0.204_{\pm0.010}$  & $0.195_{\pm 0.005}$\\
\checkmark & \checkmark & $\mathbf{0.545_{\pm0.019}}$ & $\mathbf{0.109_{\pm0.000}}$ & $\mathbf{0.157_{\pm0.001}}$ \\ \midrule
\multicolumn{2}{c}{Source} & $0.561_{\pm0.001}$ & $0.109_{\pm 0.001}$ & $0.161_{\pm 0.001}$ \\ \bottomrule
\end{tabular}}
\end{subtable}
\end{table}

\textbf{Parameter Selection}
Beyond IWV, \ac{UDA} provides several alternative strategies for model selection. A commonly used baseline is \emph{source-best} selection, in which the model with the lowest loss on source samples is chosen. Comparing these two methods in Table \ref{tab:selectionablation}, it becomes visible that IWV substantially stabilizes naive source-based selection. Especially when the gap between the distributions is not too large, source-best exhibits high variance and thereby selects optimal results. This is not the case for IWV, where only results that are on par with or better than the source model are selected. 

\begin{table}[b!]
\small
\centering
\caption{RMSE comparison of two model selection algorithms: IWV and \textit{source-best} on the SIMSHIFT dataset. Best scores are bolded.}
\label{tab:selectionablation}
\begin{subtable}{0.9\linewidth}
\centering
\resizebox{\linewidth}{!}{%
\begin{tabular}{lccc}
\toprule
Selection Method & Rolling & Motor & Forming \\ \midrule
Source Best & $0.550$ & $0.203$ & $0.157$ \\
IWV & $\mathbf{0.545}$ & $ \mathbf{0.109}$ & $\mathbf{0.157}$ \\ \bottomrule
\end{tabular}}
\end{subtable}
\end{table}

\textbf{Compute}
Compared to SSA, our method adds a moderate computational overhead.
At test-time, D-optimal source samples are forwarded through the network to estimate the density ratios used in the regularization term.
This introduces only small memory overhead, since the source samples can be fed jointly with the target batch.
The main source of additional runtime comes from the source regularization term, which increases the size of the computational graph.
Overall, we observe an approximately $1.88\times$ increase in runtime compared to SSA.
Our proposed learning rate sweeps can be executed in parallel, therefore they do not add significant runtime overhead.
\cref{tab:runtime} provides an empirical runtime comparison.

\begin{table}[ht]
\small
\centering
\caption{Runtime comparison between \ac{SSA} and \ourmethod~on the Rolling dataset, highlighting the additional overhead of the proposed method. Mean $\pm$ std across 10 runs.}
\label{tab:runtime}
\begin{subtable}{0.8\linewidth}
\centering
\resizebox{\linewidth}{!}{%
\begin{tabular}{lcc}
\toprule
\ac{TTA} Method & Runtime & Increase\\ \midrule
SSA & $0.472_{\pm 0.053}$  & \\
SATTS & $0.889_{\pm 0.085}$ & $(\uparrow\,1.88\times)$ \\ \bottomrule
\end{tabular}}
\end{subtable}
\end{table}

\section{Conclusion and Future Work}
In this work, we take an initial step toward reliable test-time adaptation for neural surrogates and, more broadly, for high-dimensional multivariate regression. Our main methodological contribution is the use of D-optimal statistics within a unified framework to stabilize test-time adaptation at three critical stages: feature alignment, regularization, and parameter tuning. The proposed adjustments enable \ac{TTA} to achieve consistent zero-shot performance improvements at negligible computational cost.

In addition to the near-zero cost gains, this line of research is particularly timely due to evolving compliance requirements.
Article 15 of the EU Artificial Intelligence Act states that high-risk AI systems need to ensure appropriate levels of accuracy and robustness \citep{EUAIAct2024Art15}.
Should neural surrogates be deployed in safety-critical domains, such as accelerating structural design in the automotive industry, accurate and reliable predictions become indispensable.

However, analyzing the \textit{``Oracle''} and \cref{fig:rel_improvement_tta} reveals clear opportunities for improvement.
This points to the potential for a new class of \ac{TTA} algorithms, specifically developed for physics simulation data. We foresee two paths to achieve \textit{``physics-driven''} \ac{TTA} that are to be explored: 
\begin{enumerate*}[label=(\roman*)]
\item use physics-informed constraints and priors \citep{raissi2019physics, cai2021physicsinformedneuralnetworkspinns}, ad-hoc and calibrated on the test case, to augment the expressiveness of the limited test labels, and
\item incorporate uncertainty quantification to localize failure regions in the fields where adaptation is necessary.
\end{enumerate*}

\section*{Acknowledgments}
We wish to thank Stephanie Holly and Florian Sestak for helpful
discussions and feedback. The ELLIS Unit Linz, the LIT AI Lab, the Institute for Machine Learning, are supported by the Federal State Upper Austria. We thank the projects FWF AIRI FG 9-N (10.55776/FG9), AI4GreenHeatingGrids (FFG- 899943), Stars4Waters (HORIZON-CL6-2021-CLIMATE-01-01), FWF Bilateral Artificial Intelligence (10.55776/COE12). We thank NXAI GmbH, Silicon Austria Labs (SAL), Merck Healthcare KGaA, GLS (Univ. Waterloo), T\"{U}V Holding GmbH, Software Competence Center Hagenberg GmbH, dSPACE GmbH, TRUMPF SE + Co. KG.

\section*{Impact Statement}
This paper presents work whose goal is to advance the field of Machine Learning applied to neural surrogates of simulations and design optimization.
There are many potential societal consequences of our work, none which we feel must be specifically highlighted here.


\bibliography{references}
\bibliographystyle{icml2026}

\newpage
\appendix
\onecolumn

\section{Supplementary Approach Information}
\label{apd:sup_approach}

\textbf{Significant-Subspace Alignment} is a TTA method for one-dimensional regression \cite{adachi_ssa_2025}. It consists of two steps: \emph{feature alignment} and \emph{significant-subspace alignment}. In the first step, source statistics such as mean $\mu^{\mathrm{src}}$ and covariance $\Sigma^{\mathrm{src}}$ are computed after source training. In the second step, a significant subspace is detected by manually selecting the top eigenvalues $\lambda_k$ of the source covariance $\Sigma^{\mathrm{src}}$. Each subspace direction $v_k^{\mathrm{src}}$ is then weighted by its influence on the regression output:
\[
{\boldsymbol{\alpha}_k = 1 + |\mathbf{w}^\top \mathbf{v}_k^{\mathrm{src}}|,}
\] 
where $\boldsymbol{\alpha}_k \geq 1$ ensures that dimensions that strongly affect the regression output are emphasized.

At test time, the precomputed source statistics are used to project the target features into the significant subspace. From the projected target features, their mean and variance $(\tilde\mu^{\mathrm{tgt}}_k, \tilde\sigma^{\mathrm{tgt}}_k{}^2)$ are calculated and aligned with the corresponding source statistics $(0, \lambda^{\mathrm{src}}_k)$. The adaptation objective is a weighted symmetric Kullback-Leibler divergence between assumed normal distributions:
\begin{equation}
\label{eq:kl_adapt}
{\mathcal{L}_{\text{TTA}} = \frac{1}{2} \sum_{k=1}^K \boldsymbol{\alpha}_k \left(\frac{(\tilde\mu^{\mathrm{tgt}}_k)^2 + \lambda^{\mathrm{src}}_k}{\tilde\sigma^{\mathrm{tgt}}_k{}^2} + \frac{(\tilde\mu^{\mathrm{tgt}}_k)^2 + \tilde\sigma^{\mathrm{tgt}}_k{}^2}{\lambda^{\mathrm{src}}_k} - 2 \right)}.
\end{equation}


\section{TTA Training}
\label{apd:tta_setup}
\textbf{Model Architecture and Representation}
In our specific setup, task-dependent parameters, such as thickness or temperature, are encoded through a conditioner network. The resulting conditioning output is passed to the base model, which, in our case, is a Transolver or Diffusion model. We extract features from the main body’s output and define the split between the representation learner and the predictor.

The exact location of this split depends on the dataset. For SIMSHIFT, the network is split before the decoder, such that the conditioner and the Transolver body together constitute the representation learner $\phi$, while the decoder acts as the predictor $g$. For EngiBench, the split is applied after the conditioner, meaning that the conditioner serves as the representation learner $\phi$ and the remaining Diffusion network functions as the predictor $g$.

\textbf{Test-Time Adaptation and Training Procedure}
For all \ac{TTA} experiments, validations source data are used to compute the statistical information $\mu_{\mathrm{src}}$ and $\sigma_{\mathrm{src}}$. In addition, a representative subset of source samples is selected from the validation set using Algorithm~\ref{alg:doptimal}, and the corresponding set is stored for training and evaluation. 

At test time, the precomputed source statistics enable the projection of the target features into the subspace. Based on the projected target features, mean and variance $(\tilde\mu^{\mathrm{tgt}}_k, \tilde\sigma^{\mathrm{tgt}}_k{}^2)$ are calculated and aligned with the corresponding source statistics $(0, \lambda^{\mathrm{src}}_k)$. We perform model updates as described in Eq.~\eqref{eq:combined_loss}. For adaptation, we only utilize the target test data, and for the regularizer, the d-optimal selected samples. We balance these losses based on the number of source samples compared to the target batch size. We chose this weighting since there is a high imbalance in information between the two losses. The amount of adaptation updates is limited by the number of available batches in each target dataset.
Adaptation is restricted to layer normalization \citep{ba2016layernormalization} parameters: for EngiBench, only the layer normalization layers of the conditioner are updated, whereas for SIMSHIFT, layer normalization parameters of both the Transolver and the conditioner are adapted. All remaining parameters are kept fixed.

For parameter tuning, we compute the latent density ratio after a single forward pass through the test-time-adapted model. To estimate this ratio, the latent source and target mean and covariance are computed and stored prior to model adaptation. These statistics are the basis for estimating the density ratio between the source and target latent distributions. Since very-high dimensional settings are prone to a lot of noise in the covariance estimation \cite{zhang2023improving}, we decided to perform a dimension reduction to improve robustness. This enables reliable covariance estimation using the D-optimally selected source samples. For each D-optimal source sample, the density ratio is computed, and the resulting values are aggregated into a loss that is used for model selection. 

As described in Section \ref{subsec:featurealign}, model selection is performed after \ac{TTA} based on IMV criterion. TThe search over learning rates ($lr$) is terminated based on performance measured on the D-optimally selected source samples. We use the Root Mean Squared Error (RMSE) for the SIMSHIFT dataset and the COMP metric for EngiBench. We set the hyperparameter search for the learning rates to $[0.05, 0.01, 0.005, 0.001, 0.0005, 0.0001]$. 

We follow standard \ac{TTA} practice and use batch size of 64 for all experiments.
To ensure robustness, we repeat each experiment with 20 random seeds per model for the SIMSHIFT benchmark, 10 for the structural beam bending dataset, and 2 for the 2D heat conduction dataset.
The varying seeds are determined by the number of data samples in each dataset. Since the 2D heat conduction dataset is small and effectively contained within a single test-time batch, increasing the number of seeds did not affect the performance of the \ac{TTA} algorithm.
This is particularly important since layer normalization is updated online, after every batch.

\textbf{Baselines}
For model comparison, we evaluate existing \ac{TTA} methods commonly used in both regression and classification tasks. For SSA as well as for Tent, we follow the procedures described in their respective method sections \cite{wang2020tent, adachi_ssa_2025}. In the implementation of SSA, the top-K eigenvalues need to be identified to compute statistics only based on a sparse set of information. We do this for each dataset. For Tent, an additional modification is required for the SIMSHIFT dataset: since entropy minimization is applied by minimizing predictive uncertainty, we train a model that explicitly predicts both mean and variance.
Additionally, we report the best-performing \ac{TTA} model on SIMSHIFT that is not selected using the IWV criterion. This result serves as a lower bound, highlighting the impact of stability-aware model selection in our approach.

\section{Experimental Setup}
\label{apd:setup}
In the following paragraphs, we detail the experimental setup, including the selected models and our training and testing strategy.

\subsection{Model Architectures \& Pretraining}
We employ different model architectures to evaluate our \ac{TTA} method. The models are based on the architectures provided in the benchmark datasets \cite{setinek2025simshift} and \cite{felten2025engibenchframeworkdatadrivenengineering}, implemented in PyTorch, and designed for conditional regression or optimization tasks. 
Node coordinates are provided as inputs and embedded using sinusoidal positional encodings. Conditioning is applied through a dedicated network that processes the simulation input parameters.

\textbfp{Conditioning Network}
The conditioner maps simulation parameters into a latent representation of dimension 8. It consists of a sinusoidal encoding, followed by a small MLP, which includes two LayerNorms to stabilize training.

\textbfp{Transolver}
The Transolver architecture \citep{pmlr-v235-wu24r} starts by encoding node coordinates using sinusoidal position embeddings, followed by an MLP that produces initial feature vectors. A learned mapping then assigns each node to a slice, enabling attention operations both within slices and between them. The processed features are passed through an MLP readout to generate the final field outputs. Two conditioning mechanisms are available: concatenating the conditioning vector with input features or applying it via DiT-based modulation across the network. Conditioning is done with the dit-based modulation \citep{peebles2023scalable}. Where a latent dimension of 128, a slice base of 32, and four attention layers are used. This results in a model with 0.57M parameters. We additionally employ a larger model with 56, 128, and 8 layers for the more complex dataset, leading to 4.07M parameters.

\textbfp{Diffusion Model}
As a diffusion model, we employ a conditional U-Net \citep{ronneberger2015unet} from Hugging Face's \texttt{diffusers} library\footnotemark.
\footnotetext{\href{https://huggingface.co/docs/diffusers/main/en/api/models\#diffusers.UNet2DConditionModel}{\texttt{UNet2DConditionModel}}}

The model works as a denoiser, taking a noisy field and a conditioning vector from the conditioning network described above and producing a noise prediction.
We summarize all hyperparameters of our diffusion model in \cref{tab:diffusion_unet_hparams}.
To train the model, we use the standard \ac{DDPM} objective of noise prediction (``$\epsilon$-prediction'') with 100 diffusion steps and a \texttt{squaredcos\_cap\_v2} beta scheduler.

\textbfp{Pretraining setup}
All unregularized baseline (\textit{``Source''}) models are pretrained using the following setup:
We use an initial learning rate of $10^{3}$ with a cosine decay scheduler and weight decay of $10^{-5}$.
Training runs for up to 500, 1500, and 3000 epochs on \emph{Beams2D}, \emph{HeatConduction2D}, and \emph{SIMSHIFT}, respectively, with early stopping if the validation loss does not improve for 500 epochs.
We enable gradient clipping and maintain an Exponential Moving Average (EMA) of the model parameters with decay 0.95.
Automatic Mixed Precision (AMP) ius enabled only for the large scale \textit{heatsink} dataset; for all others we train in \texttt{float32}.
Batch size is 64 for EngiBench baselines and 16 for SIMSHIFT baselines.

\begin{table}[h]
    \caption{Hyperparameters for our conditional diffusion U-Net. This setup leads to a model size of 17.5M parameters.}
  \label{tab:diffusion_unet_hparams}
  \centering
  \small
  \begin{tabular}{lll}
    \toprule
    \textbf{Hyperparameter} & \textbf{Value} & \textbf{HF Class Argument Name} \\
    \midrule
    Block channels (low$\to$high) & $[32, 64, 128, 256]$ & \texttt{block\_out\_channels} \\
    Layers per block & $2$ & \texttt{layers\_per\_block} \\
    Transformer layers / block & $1$ & \texttt{transformer\_layers\_per\_block} \\
    Cross-attention dim & $64$ & \texttt{cross\_attention\_dim} \\
    Only cross-attention & \texttt{True} & \texttt{only\_cross\_attention=True} \\
    Normalization groups & $16$ & \texttt{norm\_num\_groups} \\
    Activation & SiLU & \texttt{act\_fn} \\
    \bottomrule
  \end{tabular}
\end{table}

\section{Distribution Shifts for EngiBench}
\label{app:engibench_splits}
\begin{figure}[htbp]
    \centering
    \includegraphics[width=0.7\textwidth]{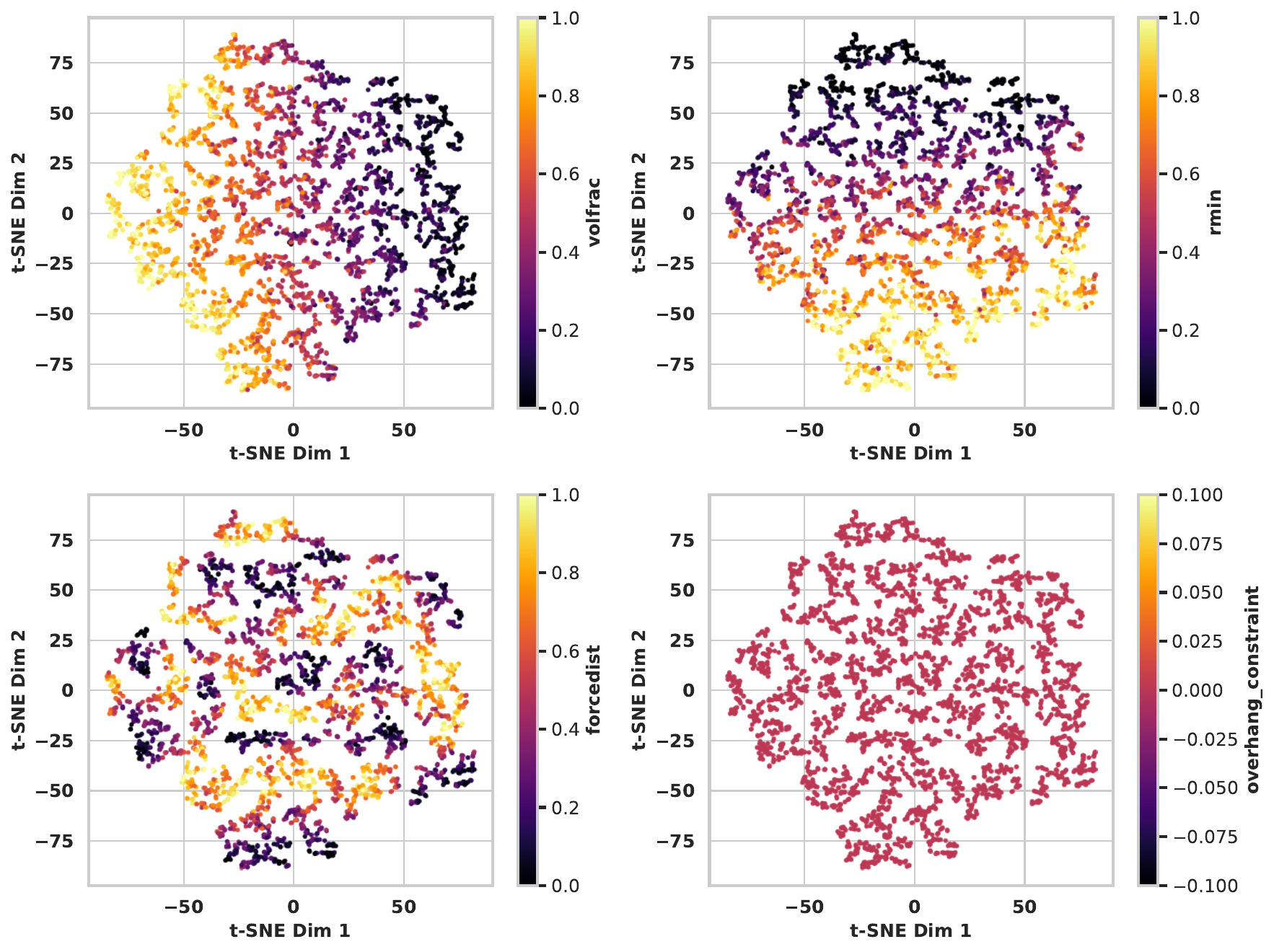}
    \caption{t-SNE visualization of the conditioner's latent space on the \emph{structural beam bending} dataset. While \texttt{overhang\_constraint} and \texttt{forcedist} are either constant or exhibit almost a uniform distribution, \texttt{volfrac} and \texttt{rmin} exhibit a clear structure.}
    \label{fig:tsne_beams}
\end{figure}

\begin{figure}[htbp]
    \centering
    \includegraphics[width=0.7\textwidth]{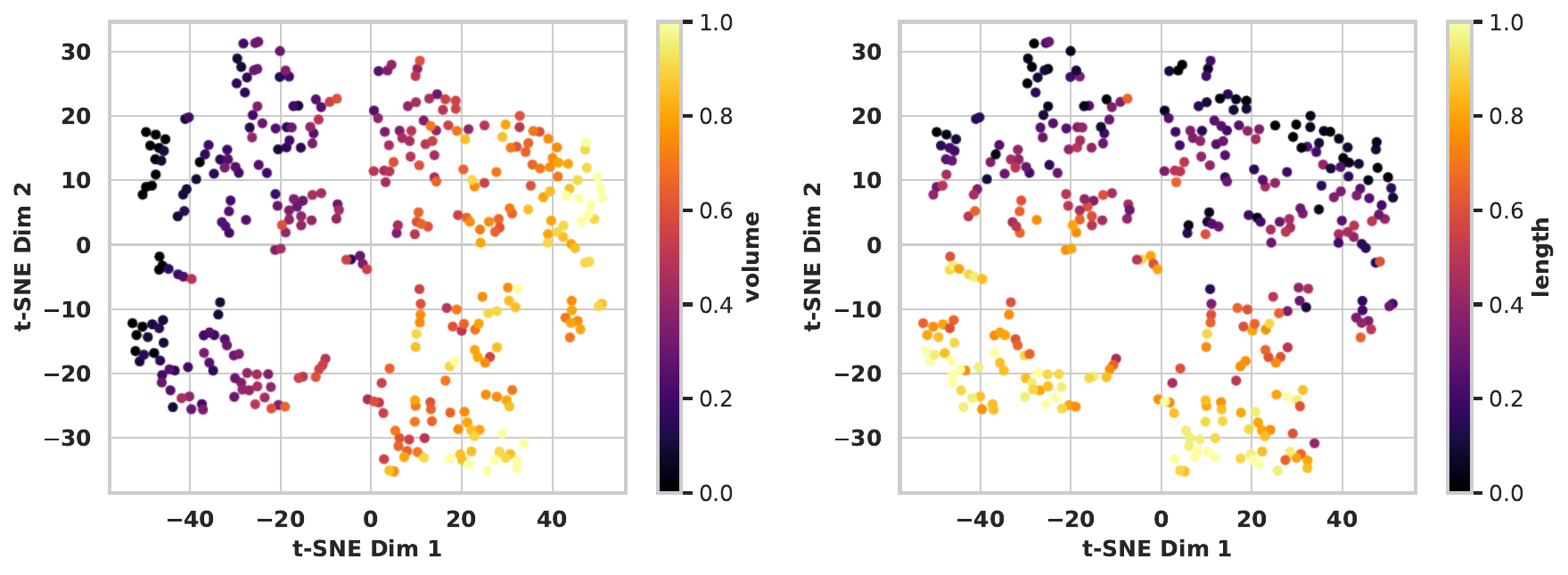}
    \caption{t-SNE visualization of the conditioner's latent space on the \emph{structural beam bending} dataset. Both conditions (\texttt{volume} and \texttt{lentgh}) exhibit a clear structure in the latent space.}
    \label{fig:tsne_heat2d}
\end{figure}

\crefrange{fig:tsne_beams}{fig:tsne_heat2d} show t-SNE visualizations of the conditioning-networks' latent spaces for models trained across the full range condition variables.
For \emph{structural beam bending} (\cref{fig:tsne_beams}), \texttt{volfrac} and \texttt{rmin} cause the clearest structure in latent space.
We therefore chose to split the source and target domain depending on the value range of \texttt{rmin}.
For \emph{2D heat conduction} (\cref{fig:tsne_heat2d}), \texttt{volume} and \texttt{length} exhibit comparable influence on the latent space distribution.
Following the same protocol, split along \texttt{volume}.
The resulting source and target ranges and sizes for both datasets can be found in \cref{table:distribution_shifts}.

\begin{table}[htbp]
\caption{Defined distribution shifts (source and target domains) for each dataset.}
\label{table:distribution_shifts}
\centering
\resizebox{0.95\textwidth}{!}{
\begin{tabular}{lllcc}
\toprule
Dataset & Parameter & Description & Source range (no.\ samples) & Target range (no.\ samples) \\
\midrule
Beams2D  & rmin   & Minimum feature length of beam members.                 & [1.5, 3.25) (3087) & [3.25, 4] (353) \\
HeatConduction2D & volume & Volume limits on the material distributions.            & [0.3, 0.465) (231) & [0.465, 0.6] (39) \\
\bottomrule
\end{tabular}
}
\end{table}





\end{document}